\documentclass[sigconf,nonacm,balance=false]{acmart}

\setcopyright{none}
\acmYear{2026}
\acmDOI{}
\acmISBN{}
\usepackage{microtype}
\usepackage{graphicx}
\usepackage{booktabs}
\usepackage{amsmath}
\usepackage{amsfonts}
\usepackage{tabularx}
\usepackage{xcolor}
\usepackage{tikz}
\usepackage{float}
\usetikzlibrary{arrows.meta,positioning}
\graphicspath{{figures/}}

\fancypagestyle{routingplain}{%
  \fancyhf{}
  \fancyfoot[C]{\thepage}
  
  }

\title{Set-Valued Routing for Language-Model Agent Systems: A WildChat Benchmark and Cost-Aware Evaluation}

\author{Ananto Nayan Bala}
\affiliation{%
  \institution{Ahsanullah University of Science and Technology}
  \city{Dhaka}
  \country{Bangladesh}}
\email{nayan.ananto@gmail.com}

\author{Faisal Muhammad Shah}
\affiliation{%
  \institution{Ahsanullah University of Science and Technology}
  \city{Dhaka}
  \country{Bangladesh}}
\email{faisal.cse@aust.edu}
\renewcommand{\shortauthors}{Ananto Nayan Bala and Faisal Muhammad Shah}

\begin{document}

\begin{abstract}
Language-model agent systems must map natural-language requests to one or more specialized agents before planning or tool execution. This decision is often treated as top-1 selection, although compositional requests can require several capabilities and unnecessary dispatch raises cost. We formulate fixed-catalog routing as set-valued language understanding and construct a benchmark of 3,000 WildChat prompts over 12 capability-defined agents. Reproducible schema-guided reference sets are balanced by agent frequency and route cardinality. We compare profile retrieval, multilabel classifiers, a fine-tuned text encoder, and catalog-conditioned zero-shot GPT-4o. Evaluation combines set quality, latency, capability coverage, and ordinal-cost utility. The encoder attains 89.59 F1, compared with 71.79 for linear SVM and 41.51 for GPT-4o, and retains 88.02 F1 under a harder semantic-group split. We further introduce Weighted Agent Routing (WAR), a deterministic post-scoring rule that changes the quality--cost operating point without retraining; Encoder+WAR obtains the best constrained utility. The results show that catalog-specific supervision and explicit set construction are important for multi-agent dispatch, while cost-aware decision layers can complement strong language encoders.
\end{abstract}

\maketitle
\pagestyle{routingplain}
\thispagestyle{routingplain}

\section{Introduction}
Language-model agents increasingly operate through catalogs of specialized capabilities: retrieval, database access, external APIs, analysis, forecasting, visualization, and report generation. Before an agentic system can plan or execute, it must decide which capabilities should receive a user's request. This routing boundary is consequential. Omitting a required agent can make a workflow impossible, while dispatching unnecessary agents increases latency, monetary cost, and the number of components exposed to untrusted input.

Many routing formulations select one tool, model, or expert. Natural-language requests, however, are often compositional: ``fetch a series, detect anomalies, and plot them'' calls for a set of capabilities. We therefore study routing as \emph{set-valued prediction}. Given a prompt and a fixed agent catalog, the router returns a small, variable-cardinality agent set. This formulation separates two questions that are often entangled in end-to-end agent systems: which agents are relevant, and how they should subsequently coordinate or execute.

The study focuses on the first question. We derive 3,000 linguistically diverse prompts from WildChat \citep{zhao2024wildchat} and map them to a 12-agent capability inventory using a documented schema-guided labeling protocol. The resulting reference routes are deliberately protocol-defined: an open-ended request may admit several reasonable decompositions under different assumptions about redundancy, cost, or desired output. The benchmark is thus intended for controlled comparison of routing policies, not as a claim of universal intent ground truth.

We compare semantic profile retrieval, classical multilabel models, a fine-tuned text encoder, and a catalog-conditioned zero-shot GPT-4o router. We also introduce Weighted Agent Routing (WAR), a deterministic decision layer that subtracts an ordinal agent-cost penalty from relevance scores before set construction. This design allows quality--cost operating points to change without retraining the underlying language model.

The paper makes four contributions:
\begin{itemize}
    \item a set-valued formulation of language-based agent routing that represents compositional requests directly;
    \item a reproducible WildChat-derived benchmark with 12 capability agents, controlled route cardinality, and balanced catalog coverage;
    \item a shared evaluation of retrieval, supervised multilabel, fine-tuned encoder, and zero-shot LLM routers using set-level, execution-oriented, and semantic-group robustness measures; and
    \item WAR, a simple cost-aware post-scoring layer, together with an analysis of how it changes routing utility for linear and encoder backbones.
\end{itemize}

Code, data, configurations, and result aggregates are available at \url{https://github.com/nayanananto/multi-agent-routing}.

\section{Related Work}
\subsection{Tool Use and Language-Model Agents}
Tool-augmented language models learn or are prompted to select APIs and external functions. Toolformer learns when and how to call tools from self-generated supervision \citep{schick2023toolformer}; ToolLLM and Gorilla study large API inventories \citep{qin2024toolllm,patil2024gorilla}; and ToolkenGPT represents tools through learned embeddings \citep{hao2023toolkengpt}. Agent frameworks such as TaskWeaver, AutoGen, MetaGPT, and ChatDev organize tools or role-specialized language models into larger workflows \citep{qiao2023taskweaver,wu2024autogen,hong2024metagpt,qian2024chatdev}. AgentBench evaluates language-model agents in interactive environments, while $\tau$-bench evaluates policy-constrained tool--agent--user interactions \citep{liu2024agentbench,yao2025taubench}. These systems motivate our setting, but commonly evaluate tool execution or complete workflows. We isolate the preceding dispatch decision and allow more than one agent to be selected.

\subsection{Routing Models and Multi-Agent Composition}
Cost-aware LLM routing usually assigns each query to one model from an ensemble. Reward-guided routing, Hybrid LLM, and FrugalGPT balance response quality against inference expense \citep{lu2024routing,ding2024hybrid,chen2023frugalgpt}. RouterBench standardizes cost--quality comparisons across multi-LLM routers, RouteLLM learns routing from preference data, and TensorOpera Router learns query-space dispatch among specialized models \citep{hu2024routerbench,ong2025routellm,stripelis2024tensoropera}. LLMRouterBench extends unified router evaluation at larger scale \citep{li2026llmrouterbench}. Output-level systems such as LLM-Blender instead rank and fuse candidate model responses \citep{jiang2023llmblender}. Our candidate items are capability-defined agents rather than interchangeable language models, and the output is a set chosen before execution rather than one selected model or a fusion of generated responses.

MasRouter addresses a broader multi-agent-system routing problem that jointly determines collaboration mode, roles, and underlying LLMs \citep{yue2025masrouter}. Our scope is complementary: we hold the catalog fixed and isolate the initial natural-language-to-capability mapping. This narrower boundary enables controlled diagnosis of under-routing, over-routing, and cost-aware set selection before planning and inter-agent communication are introduced.

\subsection{Multilabel Language Understanding}
Predicting several agents from one utterance is a multilabel text-classification problem, in which one instance may be associated with a variable-size label set \citep{zhang2014review}. Classifier Chains model label dependencies autoregressively \citep{read2011classifierchains}, while ML-kNN estimates per-label posteriors from neighboring examples \citep{zhang2007mlknn}. Related multi-intent language-understanding work jointly predicts several intents and models their interaction with slots \citep{qin2020agif}. Sentence-BERT and MPNet provide dense representations for semantic matching and classification \citep{reimers2019sentencebert,song2020mpnet}. We place these approaches under one set-construction protocol and contrast them with direct prompted-LLM selection; unlike multi-intent SLU, our labels denote executable agent capabilities rather than dialogue acts.

\section{Task and Benchmark}
\subsection{Set-Valued Agent Routing}
Let $\mathcal{A}=\{a_1,\ldots,a_M\}$ be a fixed catalog of agents and $x$ a natural-language request. A protocol-defined reference route is $G(x)\subseteq\mathcal{A}$, with $1\leq |G(x)|\leq 3$. A router produces scores $s_i(x)$ or directly emits a set $\hat{S}(x)\subseteq\mathcal{A}$. Each agent has an ordinal cost tier $c(a_i)\in\{1,2,3\}$. The goal is to cover the capabilities in $G(x)$ with a compact set while avoiding unnecessary or expensive dispatch.

The 12-agent inventory spans five access agents (time-series, SQL, API, log, and metadata retrieval), four analytic agents (statistical, trend, anomaly, and forecast), and three presentation agents (plot, summary, and report generation). Appendix~\ref{app:catalog} lists the complete inventory and cost tiers.

\subsection{WildChat-Derived Prompts}
WildChat contains naturally occurring interactions with deployed conversational models \citep{zhao2024wildchat}. We retain user prompts that activate at least one capability in the fixed inventory. The resulting benchmark contains $N=3{,}000$ unique prompts. Its natural language is inherited from WildChat, while the routing targets and class distribution are introduced by our protocol.

Reference routes are produced by reproducible schema-guided rules. Agent-specific lexical and semantic cues first identify candidate capabilities. Explicit composition patterns then retain multiple agents for requests that combine distinct operations, such as retrieval followed by analysis or forecasting followed by visualization. Sets are capped at three agents. The candidate pool is subsequently selected by a deterministic balancing procedure that enforces 1,800 single-agent, 900 two-agent, and 300 three-agent routes and 375 occurrences of every agent. We refer to these outputs as \emph{reference sets}, rather than universal gold intents, because different deployments may reasonably decompose the same request differently.

We use a fixed split stratified by reference-set size: 2,400 train, 300 development, and 300 test prompts. No normalized exact prompt occurs in more than one partition. Rebalancing makes comparisons stable across the catalog but intentionally differs from the naturally skewed WildChat distribution.

\subsection{Protocol Consistency Check}
We test whether the automatically constructed routes exhibit systematic local structure. A TF--IDF cosine graph is built over all prompts, and reference-set overlap is compared between nearest-neighbor and random prompt pairs. Rank-1 neighbors have mean set Jaccard 0.601, any-label overlap 0.732, and exact-set match 0.490, versus 0.091, 0.179, and 0.034 for random pairs. Across the top five neighbors, mean Jaccard is 0.494; prompt similarity and label-set Jaccard have Pearson correlation 0.517. This analysis measures consistency, not human validity, but shows that the protocol associates similar language with substantially more similar routes than chance.

\begin{figure*}[t]
    \centering
    \includegraphics[width=0.96\textwidth]{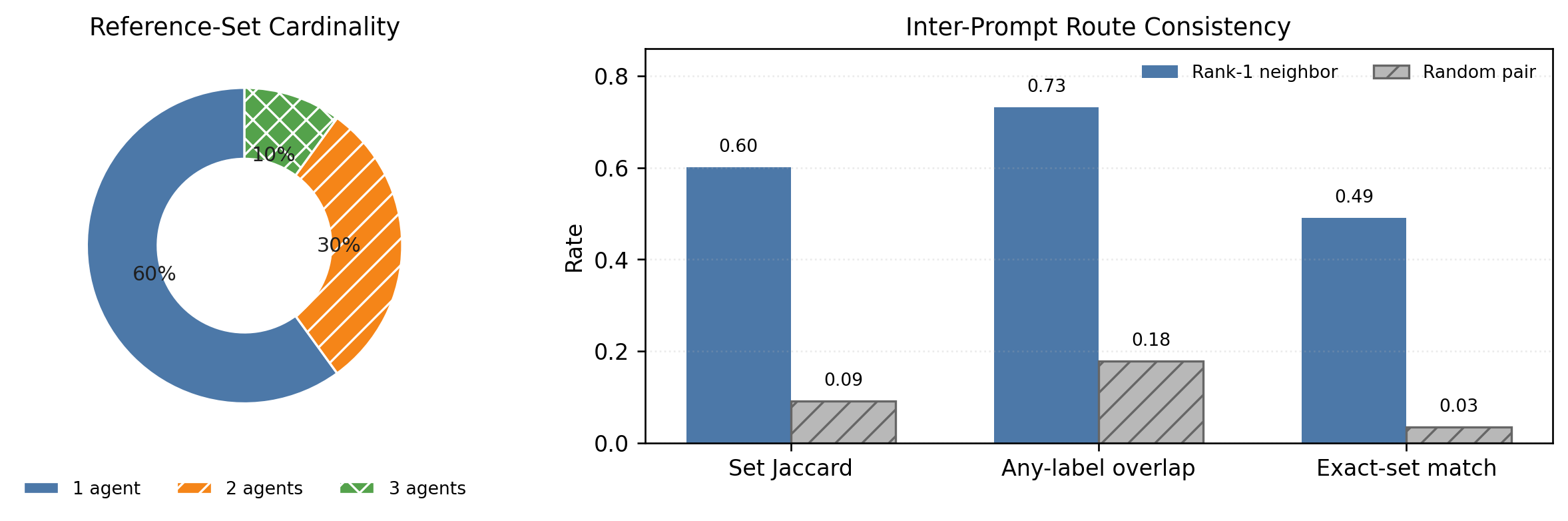}
    \caption{Benchmark construction and consistency. Left: reference-set cardinality is controlled at 60\% single-agent, 30\% two-agent, and 10\% three-agent routes. Right: rank-1 TF--IDF neighbors have substantially greater reference-set overlap than random prompt pairs. Cosine similarity and set Jaccard have Pearson correlation $r=0.517$. This is an internal consistency diagnostic, not human validation.}
    \Description{Reference routes contain one, two, or three agents in a sixty-thirty-ten percent mixture. The second panel compares rank-one neighbor overlap with random-pair overlap and reports internal consistency, not human validation.}
    \label{fig:benchmark-diagnostics}
\end{figure*}

Figure~\ref{fig:benchmark-diagnostics} summarizes both the controlled route-cardinality mix and the nearest-neighbor consistency gap.

\section{Routing Methods}
\subsection{Shared Text Representation}
The embedding-based methods encode a prompt with \texttt{all-\allowbreak mpnet-\allowbreak base-\allowbreak v2}. Let $f(x)\in\mathbb{R}^d$ denote the $\ell_2$-normalized prompt vector. Each agent also has textual profile chunks containing its role, capability description, and intent cues. All frozen-feature methods use the same MPNet encoder; the fine-tuned model starts from the same checkpoint and then adapts it to the routing catalog.

\subsection{Routing Baselines}
\paragraph{Majority.} The majority baseline always emits the most frequent agent in the training split (MetadataLookupAgent).

\paragraph{Profile retrieval (KNN).} Let $V_i$ contain the embeddings of agent $i$'s profile chunks. The retrieval score is the maximum prompt--profile cosine similarity,
\begin{equation}
s_i^{\mathrm{KNN}}(x)=\max_{v\in V_i} f(x)^\top v.
\end{equation}
This baseline requires no task-specific parameter fitting.

\paragraph{Linear SVM.} A one-vs-rest linear SVM is trained on prompt embeddings and multi-hot reference vectors. Class-balanced margins are mapped through a sigmoid to obtain per-agent scores. This model tests how much catalog-specific supervision can add without adapting the text encoder.

\paragraph{Dependency-aware models.} Classifier Chains (CC) expose earlier binary label decisions to later classifiers, while ML-kNN estimates per-agent posteriors from labeled neighbors. Both model label co-occurrence explicitly.

\paragraph{Fine-tuned encoder.} The MPNet backbone is fine-tuned jointly with a linear multilabel head. Sigmoid outputs provide one probability per agent. This model can adapt the language representation itself to the routing schema.

\paragraph{Catalog-conditioned GPT-4o.} GPT-4o receives the complete agent catalog, minimal-sufficient-set instructions, and a constrained JSON output schema, but no labeled benchmark examples. Temperature is zero, and outputs are limited to one to three valid agent names. We therefore treat it as a catalog-conditioned zero-shot baseline.

\subsection{Set Construction}
Score-producing methods use one global threshold $t$:
\begin{equation}
\hat{S}(x)=\{a_i\mid s_i(x)\geq t\}.
\label{eq:set}
\end{equation}
If the set is empty, the highest-scoring agent is selected. Development thresholds are swept over $\{0.4,0.5,0.6,0.7,0.8,0.9\}$. GPT-4o and Majority emit sets directly.

\subsection{Weighted Agent Routing}
Weighted Agent Routing (WAR) is a deterministic post-scoring policy. Given a relevance score and ordinal cost tier, it computes
\begin{equation}
\tilde{s}_i(x)=s_i(x)-\lambda c(a_i),
\label{eq:war}
\end{equation}
then applies Eq.~\ref{eq:set} to $\tilde{s}_i(x)$ with the same top-1 fallback. WAR does not retrain or replace the router; it changes the decision boundary so that a higher-cost agent needs correspondingly stronger relevance evidence. We apply WAR to the linear SVM and fine-tuned encoder. Threshold $t$ and penalty $\lambda\in\{0,0.02,0.05,0.10,0.15\}$ are jointly selected on development utility.

\begin{figure*}[t]
\centering
\includegraphics[width=0.99\textwidth]{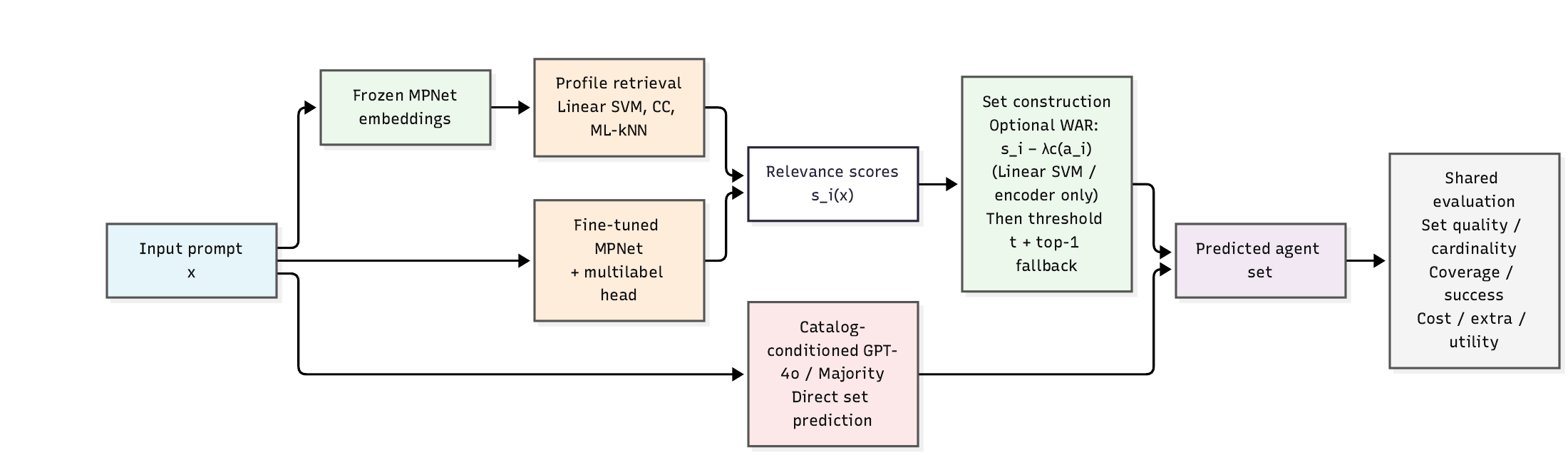}
\caption{Evaluated routing paths. Score-producing routers share thresholded set construction, with WAR optionally adjusting scores using ordinal costs. GPT-4o and Majority emit sets directly, and every predicted set is evaluated under the same protocols.}
\Description{The prompt follows frozen-MPNet scoring, a fine-tuned MPNet encoder, or a direct-set baseline. Optional WAR adjusts linear-SVM or fine-tuned-encoder scores before thresholding and top-one fallback. All predicted sets enter shared evaluation.}
\label{fig:router}
\end{figure*}

\section{Experimental Setup}
\subsection{Training and Reproducibility}
All methods use the fixed 2,400/300/300 split. The fine-tuned encoder is trained for three epochs with batch size 8, learning rate $2\times10^{-5}$, and weight decay $10^{-2}$. We report seeds 42, 43, and 44. Majority, profile retrieval, linear SVM, ML-kNN, and Linear+WAR are deterministic under the fixed split and therefore have zero variance; CC, Encoder, and Encoder+WAR vary with seed. GPT-4o is one zero-shot API run. The shared threshold $t=0.60$ is selected on development data for the unconstrained comparison. WAR configurations are selected separately by development utility and evaluated once on test data.

We additionally construct a leakage-resistant semantic-group split. A frozen multilingual MiniLM encoder embeds all prompts, and connected components in the exact cosine graph at similarity $\geq0.82$ are assigned wholly to train, development, or test. Distribution-aware assignment produces 2,400/301/299 prompts; the maximum cross-split cosine is 0.819. Linear SVM and the encoder are retrained under the same configurations, with thresholds selected on the grouped development partition.

\subsection{Set and Execution-Oriented Metrics}
For each prompt we compute sample-averaged precision, recall, F1, Jaccard similarity, exact set match (EM), and predicted set size. These distinguish missing required capabilities from adding unnecessary ones.

We also simulate the dispatch boundary without executing downstream agents. Success is $\mathbf{1}[G(x)\subseteq\hat{S}(x)]$, and capability coverage is
\begin{equation}
\mathrm{Coverage}(x)=\frac{|G(x)\cap\hat{S}(x)|}{|G(x)|}.
\end{equation}
The ordinal dispatch cost is $\sum_{a\in\hat{S}(x)}c(a)$ and $\mathrm{Extra}(x)=|\hat{S}(x)\setminus G(x)|$. Utility is
\begin{equation}
U=\mathrm{Coverage}-0.10\,\mathrm{Cost}-0.05\,\mathrm{Extra}.
\label{eq:utility}
\end{equation}
The coefficients define one transparent operating environment rather than universal monetary values. Per-query latency comes from a representative run and includes API round-trip time for GPT-4o.

\section{Results}
\subsection{Set-Valued Routing Quality}
Table~\ref{tab:main} shows a clear separation between catalog-supervised routing and direct profile or prompted-LLM selection. The fine-tuned encoder achieves the best unconstrained result: F1 $89.59\pm0.33$, Jaccard $85.52\pm0.30$, and EM $73.11\pm0.19$. Linear SVM reaches 71.79 F1 and exactly matches the average reference cardinality of 1.50. It substantially exceeds profile retrieval (42.56 F1), showing that the frozen embedding already contains useful semantic information but requires catalog-specific decision boundaries.

CC and ML-kNN occupy an intermediate range. CC favors recall and larger sets, whereas ML-kNN favors precision. Their 65.13 and 63.10 F1 scores indicate that label-dependency modeling helps relative to direct profile similarity, but representation adaptation remains more consequential in this benchmark.

Catalog-conditioned zero-shot GPT-4o obtains 41.51 F1 despite receiving full agent descriptions and a constrained output schema. Its average set size (1.40) is close to the reference average, so the gap is not explained only by over- or under-selection. Instead, the model often chooses a plausible capability set that does not match the benchmark's catalog-specific decomposition. This result cautions against assuming that broad instruction-following ability alone yields calibrated fixed-inventory routing.

\begin{table*}[t]
\centering
\small
\setlength{\tabcolsep}{5.0pt}
\caption{Test-set routing results at the shared unconstrained threshold $t=0.60$. Values are percentages except average set size. CC and Encoder report mean $\pm$ standard deviation over three seeds; deterministic methods have identical values across seeds. GPT-4o is one zero-shot run.}
\label{tab:main}
\begin{tabular}{lrrrrrr}
\toprule
Method & Precision & Recall & F1 & Jaccard & EM & Avg. $|\hat{S}|$ \\
\midrule
Majority & 10.33 & 6.94 & 7.94 & 6.94 & 4.33 & 1.00 \\
Profile retrieval & 52.00 & 38.39 & 42.56 & 38.39 & 27.00 & 1.00 \\
Linear SVM & 73.03 & 75.33 & 71.79 & 66.36 & 49.67 & 1.50 \\
Classifier Chains & $64.49\pm1.17$ & $71.96\pm0.61$ & $65.13\pm1.04$ & $58.00\pm1.26$ & $37.56\pm1.71$ & $1.65\pm0.01$ \\
ML-kNN & 71.44 & 60.50 & 63.10 & 58.36 & 45.33 & 1.15 \\
GPT-4o (zero-shot) & 46.22 & 42.33 & 41.51 & 35.56 & 19.00 & 1.40 \\
Fine-tuned Encoder & $\mathbf{97.80\pm0.23}$ & $\mathbf{85.98\pm0.42}$ & $\mathbf{89.59\pm0.33}$ & $\mathbf{85.52\pm0.30}$ & $\mathbf{73.11\pm0.19}$ & $1.20\pm0.01$ \\
\bottomrule
\end{tabular}
\end{table*}

\begin{figure}[t]
    \centering
    \includegraphics[width=\columnwidth]{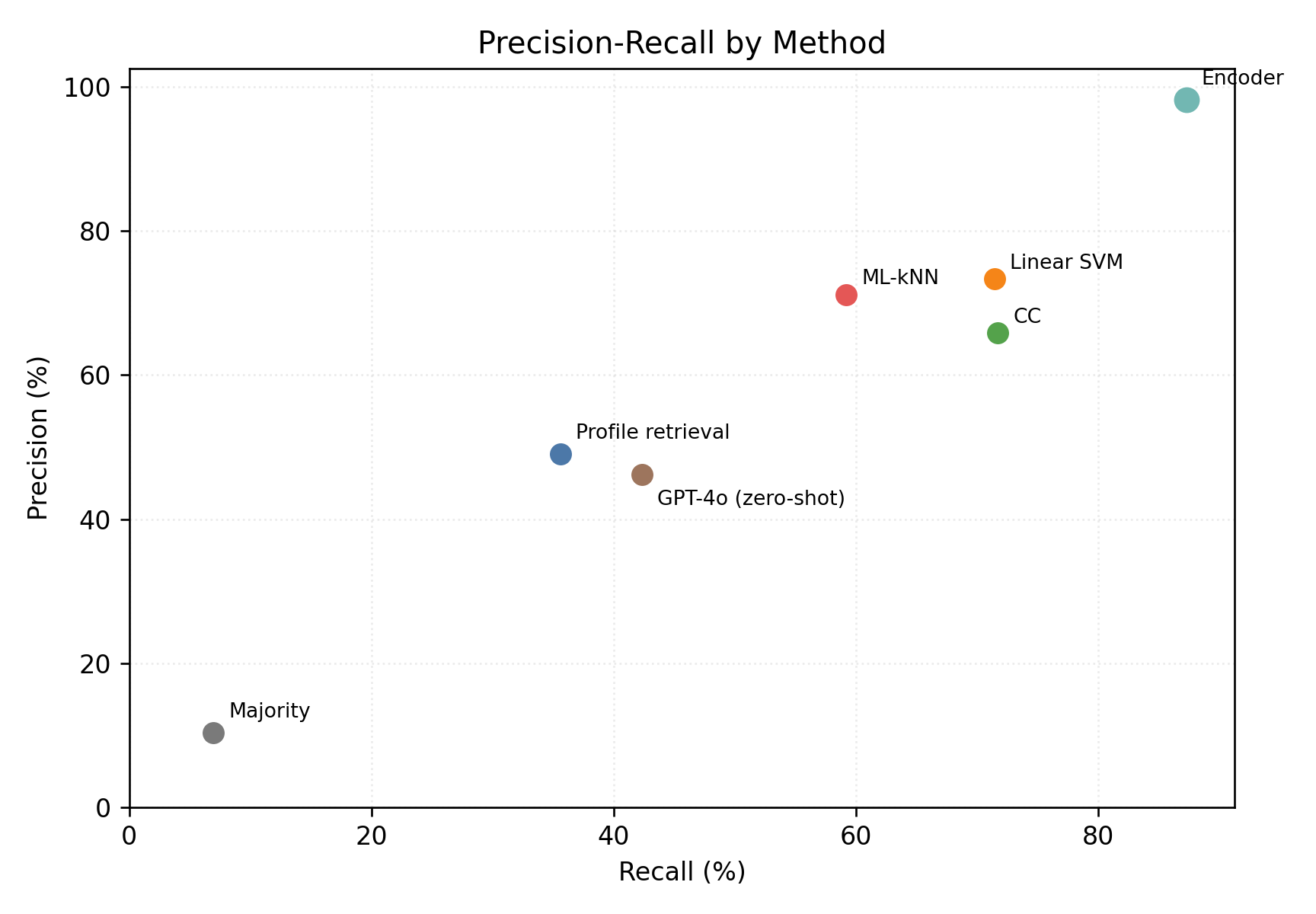}
    \caption{Precision--recall positions under the selected unconstrained protocol. GPT-4o is one zero-shot run; non-LLM results use the three-seed aggregate where applicable.}
    \Description{A precision-recall scatter plot comparing all seven routing baselines. The fine-tuned encoder has the strongest joint precision-recall position under the selected protocol.}
    \label{fig:pr}
\end{figure}

Figure~\ref{fig:pr} visualizes the precision--recall separation behind Table~\ref{tab:main}. The encoder occupies the strongest joint operating region, while the zero-shot LLM remains close to profile retrieval despite producing a comparably sized route set.

\subsection{Threshold Behavior}
\begin{figure}[t]
    \centering
    \includegraphics[width=\columnwidth]{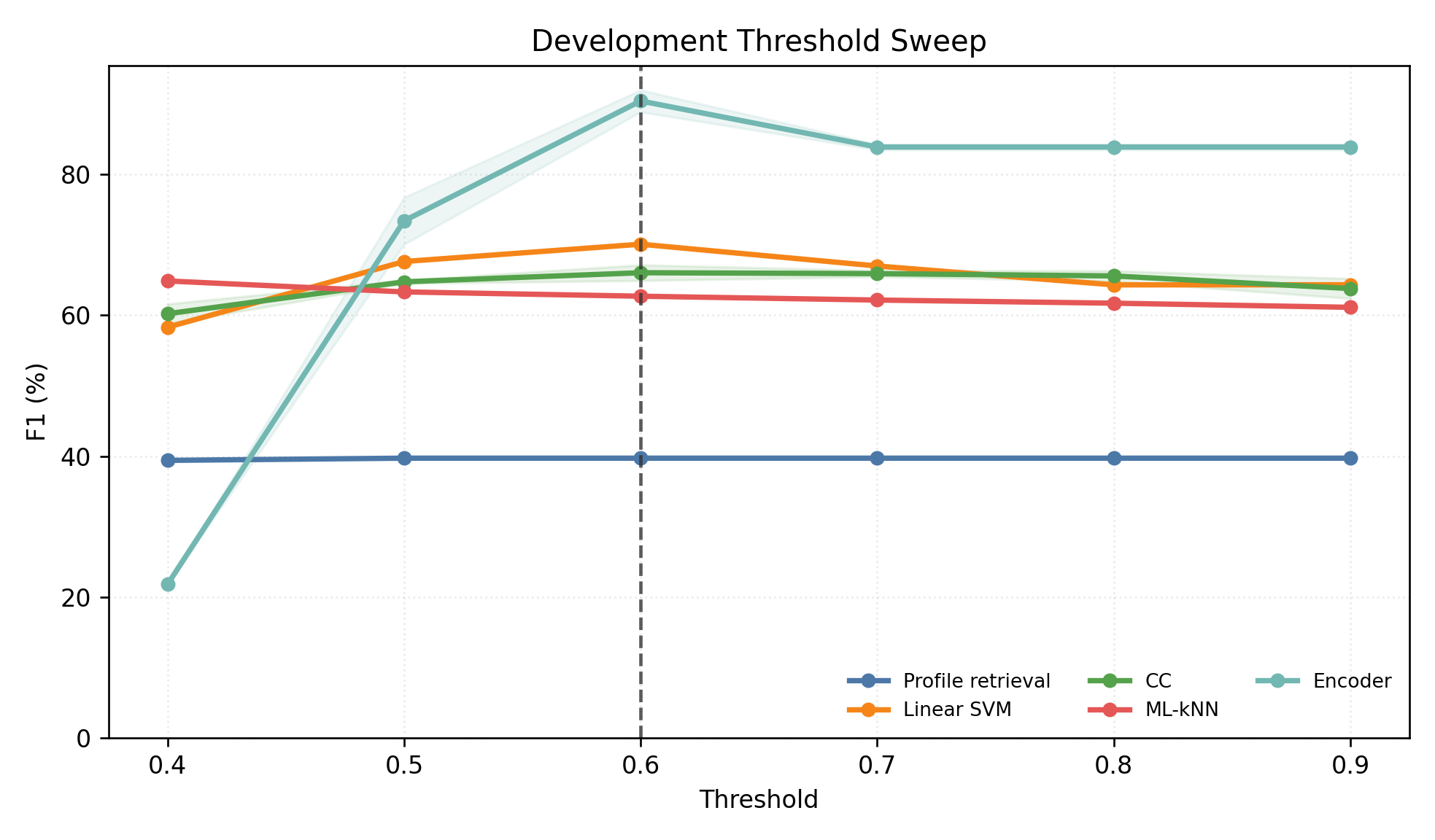}
    \caption{Development threshold sweep (mean $\pm$ standard deviation over three seeds). Lower thresholds over-select; higher thresholds suppress recall.}
    \Description{Development F1 curves as the global threshold changes. Linear SVM and the encoder peak at a threshold of zero point six under the reported sweep.}
    \label{fig:threshold}
\end{figure}

Figure~\ref{fig:threshold} shows the expected set-construction trade-off. Linear SVM rises from 58.31 F1 at $t=0.40$ to a development peak of 70.09 at $t=0.60$, then loses recall. The encoder rises more sharply, from 73.41 at $t=0.50$ to 90.42 at $t=0.60$, before falling to 83.89 at $t\geq0.70$. The shared interior optimum indicates that explicit set calibration matters even when the underlying semantic scorer is strong.

\subsection{Semantic-Group Robustness}
Table~\ref{tab:grouped} tests whether the main ranking survives after close semantic neighborhoods are kept within one partition. The split is substantially harder for the frozen-feature linear model, whose F1 falls by 10.01 points. In contrast, the fine-tuned encoder falls by only 1.57 points and retains 88.02 F1 and 69.23 exact match. The encoder's advantage therefore persists when no prompt pair with cosine similarity at or above 0.82 crosses partitions, reducing the likelihood that its primary result is explained by close train--test paraphrases alone.

\begin{table}[t]
\centering
\small
\setlength{\tabcolsep}{2.5pt}
\caption{Routing robustness under the semantic-group split. Values are percentages; encoder results are mean $\pm$ standard deviation over three seeds. ``Standard'' denotes the primary stratified split.}
\label{tab:grouped}
\begin{tabular}{@{}lrrr@{}}
\toprule
Method & Std. F1 & Grouped F1 & Grouped EM \\
\midrule
Linear & 71.79 & 61.78 & 45.15 \\
Encoder & $89.59\pm0.33$ & $88.02\pm0.92$ & $69.23\pm1.53$ \\
\bottomrule
\end{tabular}
\end{table}

\subsection{Capability Coverage, Latency, and Utility}
Table~\ref{tab:execution} asks whether a predicted set covers all reference capabilities. The encoder reaches $74.33\pm1.20$ success and $86.07\pm0.68$ coverage while adding only $0.03$ extra agents on average. Linear SVM is the strongest non-neural local alternative at 0.465 utility. The GPT-4o baseline is both less accurate and substantially slower in this representative API-backed run; latency should be interpreted as a deployment measurement rather than an intrinsic model comparison.

\begin{table*}[t]
\centering
\small
\setlength{\tabcolsep}{3.0pt}
\caption{Execution-oriented test results on 300 prompts. F1, success, and coverage are percentages; latency is milliseconds per query. Encoder rows report three-seed mean $\pm$ standard deviation. A dash indicates that separate post-processing latency was not measured.}
\label{tab:execution}
\begin{tabular}{lrrrrrrr}
\toprule
Method & F1 & Success & Coverage & Cost & Extra & Utility & ms \\
\midrule
Profile retrieval & 42.56 & 27.00 & 38.39 & 1.81 & 0.48 & 0.179 & 66.07 \\
Linear SVM & 71.79 & 66.67 & 75.33 & 2.67 & 0.43 & 0.465 & 60.94 \\
Linear+WAR & 69.72 & 66.67 & 75.39 & 2.48 & 0.53 & 0.480 & -- \\
Encoder & $89.59\pm0.33$ & $74.33\pm1.20$ & $86.07\pm0.68$ & $2.06\pm0.02$ & $0.03\pm0.00$ & $0.654\pm0.006$ & 49.93 \\
Encoder+WAR & $\mathbf{91.36\pm0.82}$ & $\mathbf{92.11\pm0.69}$ & $\mathbf{96.39\pm0.34}$ & $2.82\pm0.03$ & $0.24\pm0.03$ & $\mathbf{0.670\pm0.005}$ & -- \\
GPT-4o (zero-shot) & 41.51 & 29.33 & 42.33 & 2.01 & 0.78 & 0.183 & 1979.06 \\
\bottomrule
\end{tabular}
\end{table*}

\subsection{Cost-Aware Operating Points}
WAR changes the routing policy rather than the relevance model. To separate its contribution from ordinary threshold tuning, we also select the no-WAR ($\lambda=0$) threshold directly by development utility. This independently selects $t=0.60$ for both backbones, so the Linear and Encoder rows in Table~\ref{tab:execution} are also utility-optimized no-penalty baselines. Against these controls, Linear+WAR raises test utility from 0.465 to 0.480: average cost falls from 2.67 to 2.48 while success and coverage remain nearly unchanged. Encoder+WAR raises coverage from $86.07$ to $96.39$ and success from $74.33$ to $92.11$; the coverage gain outweighs higher cost and extra-agent penalties, increasing utility from 0.654 to 0.670. Thus, the gain is not obtained by comparing WAR with an F1-only threshold, and ``cost-aware'' denotes optimization of the stated quality--cost objective rather than cost minimization alone.

The selected development operating point uses $(t,\lambda)=(0.40,0.10)$ for Linear+WAR and $(0.50,0.02)$ for Encoder+WAR. The stronger encoder needs only a light cost adjustment, while the linear scorer benefits from a larger penalty. These are joint development-selected operating points; they should not be interpreted as universal cost settings.

\begin{figure}[t]
    \centering
    \includegraphics[width=\columnwidth]{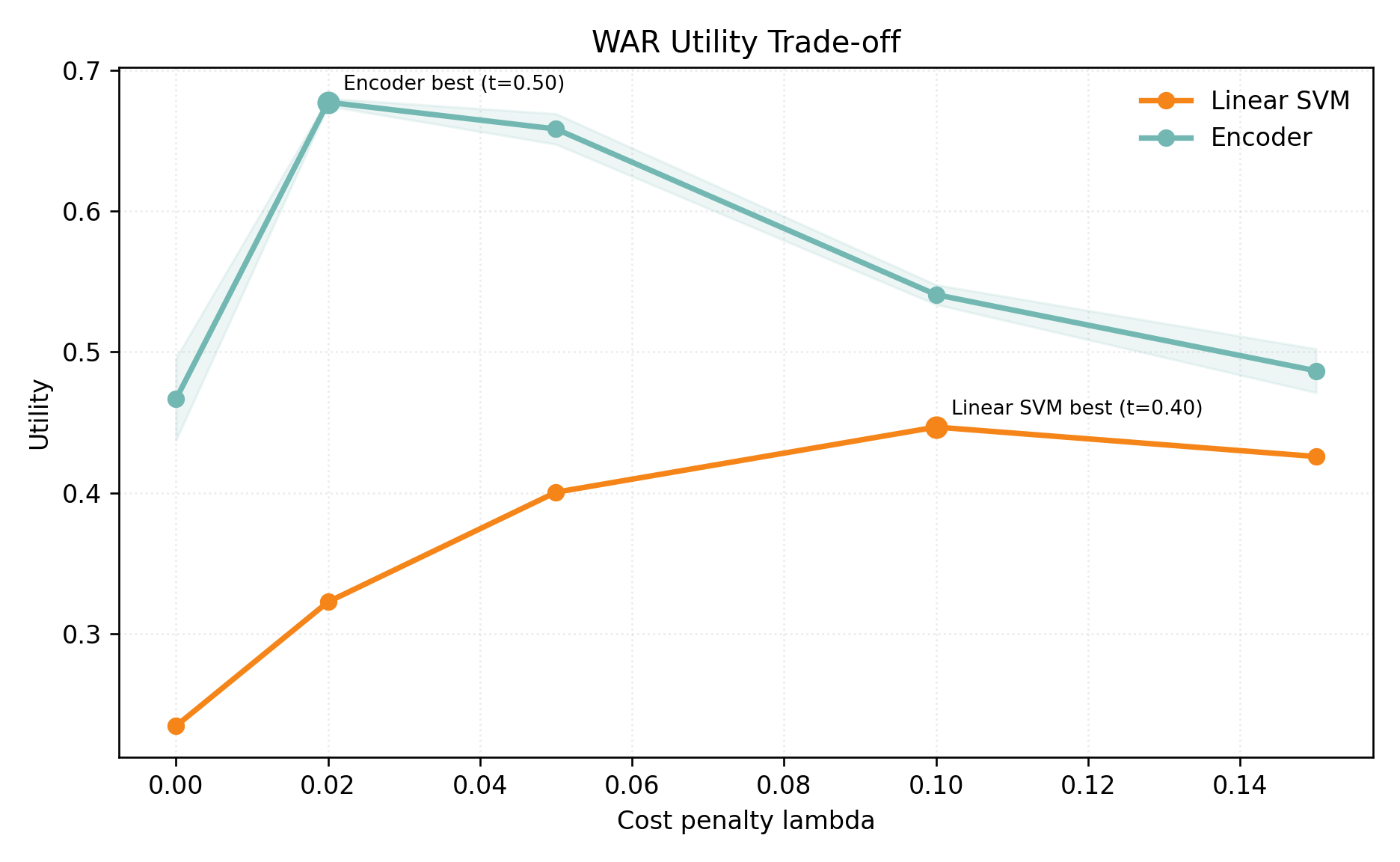}
    \caption{Development utility as the WAR penalty varies at each method's selected threshold. Labeled markers indicate the chosen operating points.}
    \Description{Development utility versus the ordinal-cost penalty for linear and encoder backbones, with their selected operating points marked.}
    \label{fig:war}
\end{figure}

\section{Discussion}
Three findings are most relevant to language-model agent systems. First, routing is not solved simply by placing a capable LLM in front of an agent catalog. The zero-shot GPT-4o router produces reasonable-size sets but does not reproduce the fixed routing protocol reliably. In contrast, even a linear classifier trained on frozen text embeddings gains roughly 30 F1 points over GPT-4o. Catalog-specific supervision therefore matters when agent definitions are operational and route cardinality must be calibrated.

Second, adapting the language representation matters more than adding label-dependency structure alone. CC and ML-kNN improve on profile retrieval, yet remain far behind the fine-tuned encoder. The semantic-group split sharpens this finding: linear SVM loses 10.01 F1 points, whereas the adapted encoder loses only 1.57. Much of the difficulty therefore lies in mapping heterogeneous wording into the local capability schema rather than recovering only close paraphrase neighborhoods.

Third, relevance estimation and deployment policy are usefully separable. WAR leaves the backbone unchanged and adjusts only the set decision. On the linear model it primarily reduces tier cost; on the encoder it chooses a higher-coverage operating point whose added cost is justified by the utility. This separation is attractive for agent platforms because cost policies can change without repeating encoder training.

The study intentionally stops at dispatch. It does not claim that selecting the reference set guarantees a correct final answer, successful coordination, or safe tool execution. Rather, it supplies a controlled diagnostic for one failure surface within a larger agent architecture. End-to-end agent benchmarks and this routing benchmark answer different questions: the former measure complete behavior, while the latter attributes errors specifically to capability selection.

\section{Conclusion}
We framed fixed-catalog language-model agent routing as set-valued prediction and evaluated it on 3,000 WildChat-derived prompts over 12 capabilities. Under a shared protocol, a fine-tuned encoder produced the most accurate agent sets, retained 88.02 F1 on a leakage-resistant semantic-group split, and substantially exceeded both linear SVM and catalog-conditioned zero-shot GPT-4o. The WAR decision layer further separated relevance scoring from a development-selected quality--cost policy and improved over utility-optimized no-WAR controls. These findings support explicit set construction and catalog-specific learning at the dispatch boundary of multi-agent systems. Future work should replace protocol-defined routes with independently annotated alternatives, evaluate natural catalog frequencies, and connect routing decisions to real multi-agent execution outcomes.

\section{Limitations}
The reference routes are produced automatically from a fixed capability schema rather than independently annotated by multiple humans. They are reproducible and internally systematic, but they can encode lexical biases and cannot establish a uniquely correct decomposition for every open-ended prompt. Inter-prompt consistency is evidence of regularity, not a substitute for human validity or inter-annotator agreement.

Balancing agent frequencies and route cardinalities improves diagnostic coverage but changes the natural WildChat distribution. The primary random stratified split has no exact duplicate prompts across partitions. Our additional semantic-group split keeps every pair with frozen-encoder cosine similarity $\geq0.82$ in one partition, but weaker semantic overlap can remain below that threshold, and neither split reproduces WildChat's natural route frequencies. Temporal and cross-catalog evaluations would provide further generalization tests.

The catalog contains only 12 English-described capabilities, and results may not transfer to larger, multilingual, dynamically changing, or more overlapping inventories. GPT-4o is evaluated in one zero-shot run, so the result does not characterize other proprietary or open models, alternative prompts, or sampling settings.

Cost tiers are ordinal proxies rather than measured prices, energy use, or context-dependent latency. The execution-oriented study simulates capability coverage and does not execute agents, evaluate ordering, model inter-agent communication, or measure final answer quality. Consequently, the reported utility identifies operating points under one transparent proxy environment rather than deployment-optimal policies.

\section{Ethical Considerations}
The benchmark derives prompts from the public WildChat research corpus. Public availability does not eliminate privacy concerns in conversational logs: prompts may contain personal, sensitive, offensive, or copyrighted material. Any redistribution must follow the source dataset's license and access conditions, preserve applicable filtering, and avoid using metadata to identify or profile users. Country and source-model fields are not used as router inputs in our experiments.

Automatically assigned reference routes can reproduce biases in the capability definitions and lexical rules. They should not be treated as human intent labels or used directly for high-stakes dispatch. In deployed systems, routing errors can omit necessary capabilities or invoke unnecessary external tools, increasing cost and attack surface. Appropriate safeguards include authorization at tool boundaries, confidence-aware fallback, rate limits on fan-out, audit logs, and human review for consequential requests.

\bibliographystyle{ACM-Reference-Format}
\bibliography{references}

\clearpage
\onecolumn
\appendix
\section{Agent Catalog}
\label{app:catalog}
\begin{table}[H]
\centering
\small
\setlength{\tabcolsep}{2pt}
\caption{Fixed 12-agent inventory. Tiers are ordinal deployment proxies: 1 denotes retrieval or light synthesis, 2 analysis or presentation, and 3 external access or inference-heavy forecasting.}
\label{tab:catalog}
\begin{tabular}{@{}p{0.34\columnwidth}cp{0.52\columnwidth}@{}}
\toprule
Agent & Tier & Capability \\
\midrule
TimeSeriesQueryAgent & 1 & Retrieve time-series observations over a requested interval. \\
SQLQueryAgent & 1 & Filter, join, group, and aggregate structured tables. \\
APIDataFetchAgent & 3 & Fetch live or recent data from external APIs. \\
LogRetrievalAgent & 1 & Retrieve and filter application or system logs. \\
MetadataLookupAgent & 1 & Find schemas, columns, endpoints, and dataset metadata. \\
StatisticalAnalysisAgent & 2 & Compute descriptive statistics, distributions, and correlations. \\
TrendAnalysisAgent & 2 & Identify temporal trends, seasonality, and direction changes. \\
AnomalyDetectionAgent & 2 & Detect spikes, drops, outliers, and unusual patterns. \\
ForecastAgent & 3 & Predict future values from historical observations. \\
PlotGenerationAgent & 2 & Produce charts and plots from numeric or temporal data. \\
SummaryAgent & 1 & Produce concise natural-language summaries. \\
ReportWriterAgent & 2 & Produce structured, detailed reports for stakeholders. \\
\bottomrule
\end{tabular}
\end{table}

\section{Benchmark Examples and Distributions}
\begin{table}[H]
\centering
\small
\setlength{\tabcolsep}{3pt}
\caption{Illustrative benchmark prompts and protocol-defined reference routes.}
\begin{tabular}{@{}p{0.61\columnwidth}p{0.31\columnwidth}@{}}
\toprule
Prompt & Reference agents \\
\midrule
Plot monthly sales trends for 2023 and highlight anomalies. & Trend, Plot, Anomaly \\
Fetch the latest CPI series and compute year-over-year change. & API Fetch, Statistics \\
Find the shipments schema and query delayed shipments last quarter. & Metadata, SQL \\
Forecast next-month demand using historical orders. & Forecast, Statistics \\
\bottomrule
\end{tabular}
\end{table}

\begin{figure}[H]
    \centering
    \includegraphics[width=0.76\textwidth]{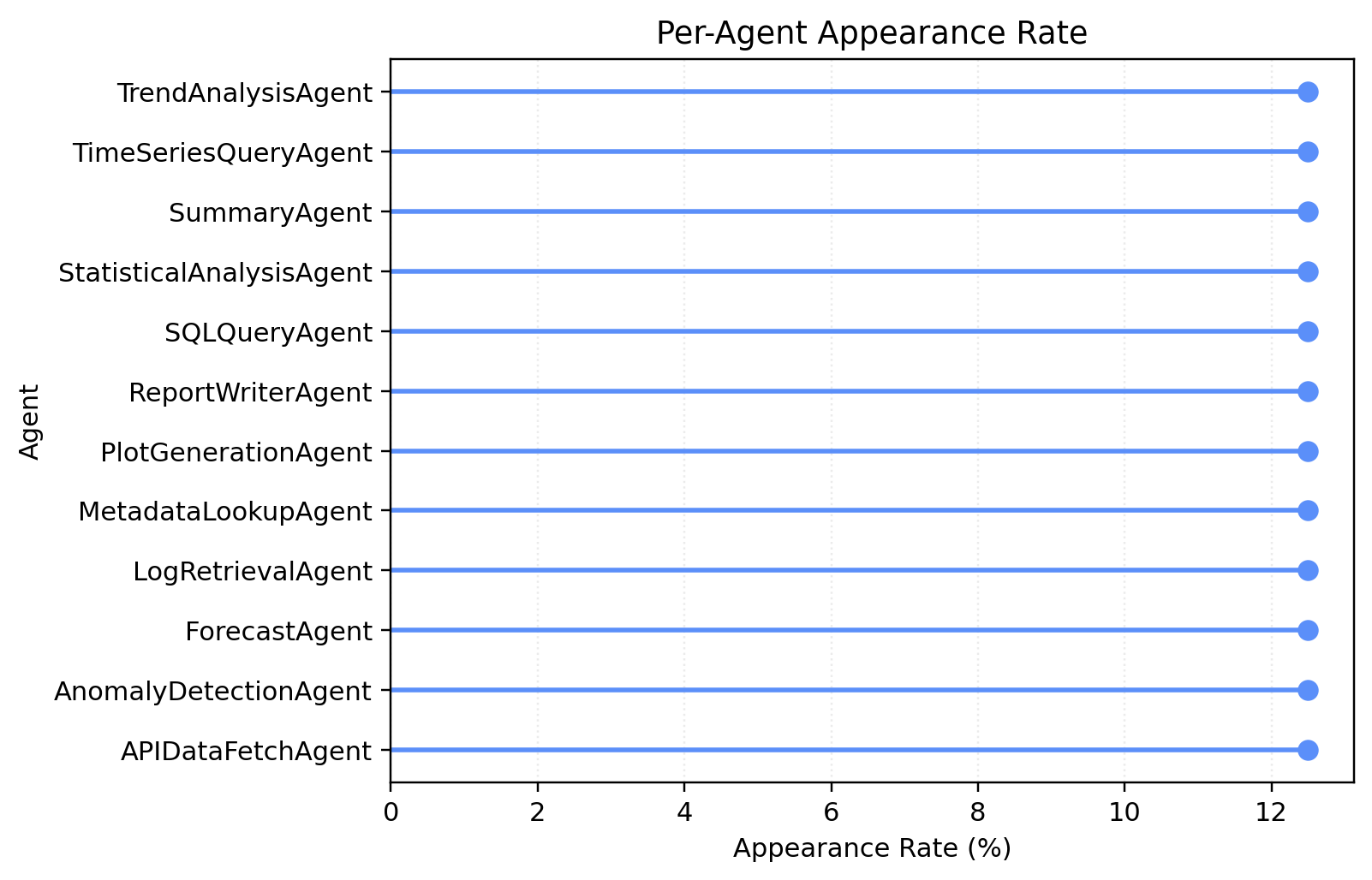}
    \caption{Per-agent appearance rates in the protocol-defined benchmark. Each capability occurs in 375 of the 3,000 prompts (12.5\%). A prompt may contain several agents, so these rates are not a partition of prompts. This controlled distribution is not the natural WildChat frequency distribution.}
    \Description{A horizontal lollipop chart showing the same 12.5 percent appearance rate for all twelve capability-defined agents.}
    \label{fig:agent-appearance}
\end{figure}

\clearpage
\section{Additional Routing View}
\begin{figure}[H]
    \centering
    \includegraphics[width=0.62\columnwidth]{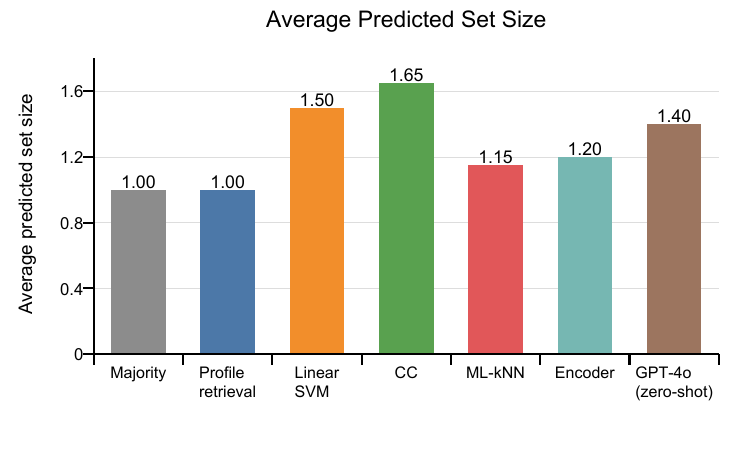}
    \caption{Average predicted set size under the selected unconstrained protocol. WAR variants are analyzed separately because they optimize execution-oriented utility.}
    \Description{Average predicted route cardinality for the seven unconstrained routing methods, ranging from one for Majority and profile retrieval to about one point six five for Classifier Chains.}
    \label{fig:setsize}
\end{figure}

\section{Deployment Context}
\begin{figure}[H]
    \centering
    \includegraphics[width=0.76\textwidth]{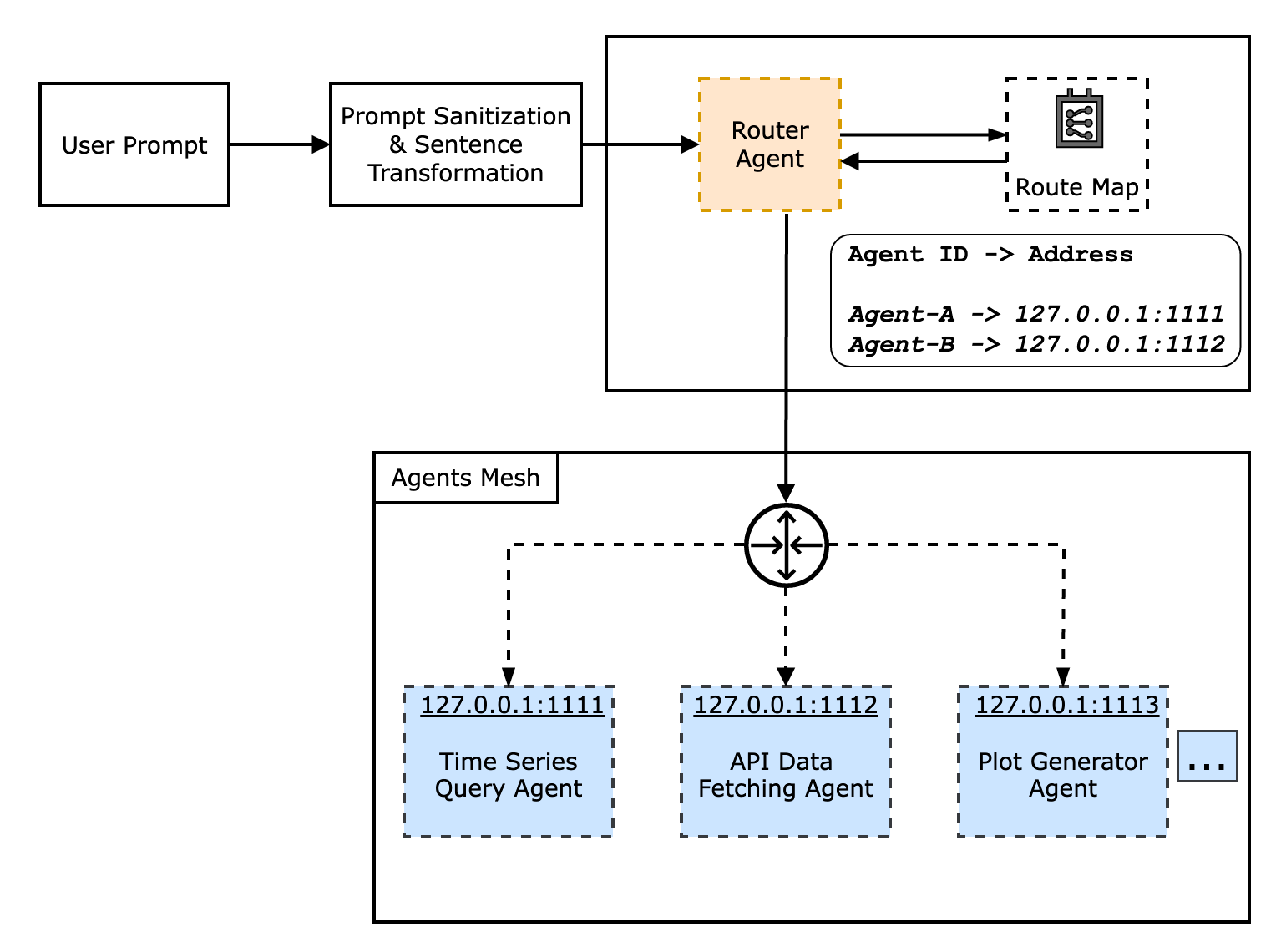}
    \caption{Illustrative deployment context from the earlier manuscript. The benchmark in this paper evaluates capability-set selection only. Prompt sanitization, route-map lookup, endpoint addresses, and downstream agent execution are shown as possible surrounding deployment components; they are not evaluated or claimed as experimental results here. Endpoint addresses are illustrative.}
    \Description{A prompt enters a router, whose selected agent identifiers could be mapped to illustrative local endpoints in a downstream agent mesh. Only the selection decision is part of the reported routing benchmark.}
    \label{fig:deployment-context}
\end{figure}

\end{document}